\documentclass[10pt,twocolumn,letterpaper]{article}

\usepackage{wacv}
\usepackage{times}
\usepackage{epsfig}
\usepackage{graphicx}
\usepackage{amsmath}
\usepackage{amssymb}

\usepackage{enumitem}
\usepackage{graphicx}
\usepackage{subfloat}
\usepackage{tabularx}
\usepackage{adjustbox}
\usepackage{amssymb}
\usepackage{multirow}
\usepackage[flushleft]{threeparttable}
\usepackage{subfigure}

\usepackage{algorithm}
\usepackage{algorithmic}

\newcommand{\figref}[1]{Fig.~\ref{#1}}
\newcommand{\tabref}[1]{Table.~\ref{#1}}
\newcommand{\eref}[1]{Eq.~(\ref{#1})}
\newcommand{\sref}[1]{Sec.~\ref{#1}}
\renewcommand{\ie}{\textit{i.e.}}

\def\wacvPaperID{1301} 

\wacvfinalcopy 

\usepackage[pagebackref=true,breaklinks=true,letterpaper=true,colorlinks,bookmarks=false]{hyperref}

\ifwacvfinal\fi
\begin{document}

\title{The Devil is in the Boundary: Exploiting Boundary Representation for Basis-based Instance Segmentation}

\author{Myungchul Kim \\
KAIST\\
{\tt\small gritycda@kaist.ac.kr}
\and
Sanghyun Woo \\
KAIST\\
{\tt\small shwoo93@kaist.ac.kr}
\and
Dahun Kim \\
KAIST\\
{\tt\small mcahny@kaist.ac.kr}
\and
In So Kweon \\
KAIST\\
{\tt\small iskweon@kaist.ac.kr}
}
\maketitle

\thispagestyle{empty}

\begin{abstract}
Pursuing a more coherent scene understanding towards real-time vision applications, single-stage instance segmentation has recently gained popularity, achieving a simpler and more efficient design than its two-stage counterparts. 
Besides, its global mask representation often leads to superior accuracy to the two-stage Mask R-CNN which has been dominant thus far. 
Despite the promising advances in single-stage methods, finer delineation of instance boundaries still remains unexcavated. Indeed, boundary information provides a strong shape representation that can operate in synergy with the fully-convolutional mask features of the single-stage segmenter. 
In this work, we propose Boundary Basis based Instance Segmentation(\textbf{B2Inst}) to learn a global boundary representation that can complement existing global-mask-based methods that are often lacking high-frequency details. 
Besides, we devise a unified quality measure of both mask and boundary and introduce a network block that learns to score the per-instance predictions of itself. 
When applied to the strongest baselines in single-stage instance segmentation, our \textit{B2Inst} leads to consistent improvements and accurately parse out the instance boundaries in a scene. 
Regardless of being single-stage or two-stage frameworks, we outperform the existing state-of-the-art methods on the COCO dataset with the same ResNet-50 and ResNet-101 backbones.
\end{abstract}

\section{Introduction}

\begin{figure}[t]
\small
\begin{tabular}{@{}c @{\hskip 0.005\linewidth}c @{\hskip 0.005\linewidth}c}
\includegraphics[width=0.325\linewidth]{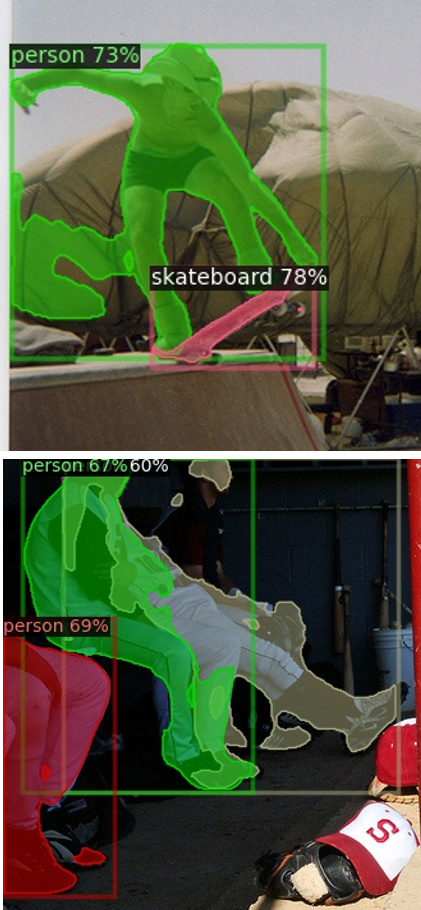} &
\includegraphics[width=0.325\linewidth]{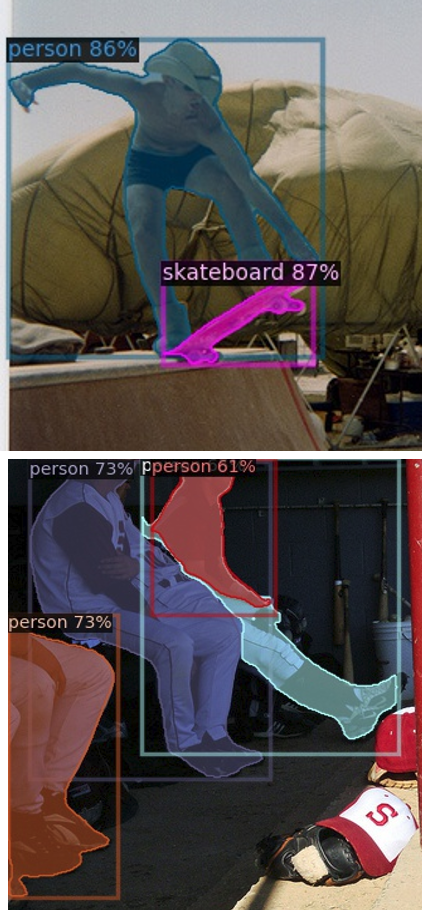} &
\includegraphics[width=0.325\linewidth]{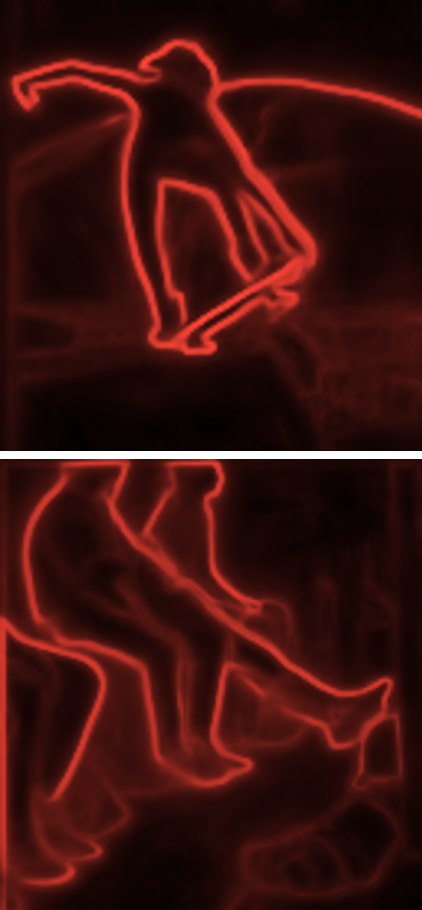} \\
(a) \footnotesize BlendMask & 
(b) \footnotesize Our B2Inst & 
(c) \footnotesize Our boundary basis \\
\end{tabular}
\vspace{2mm}
\caption{{\bf BlendMask vs B2Inst}:
    (a) Coarse boundaries appeared in the instance segmentation results of BlendMask~\cite{chen2020blendmask}. (b) Our proposed boundary-aware masks are more precise in delineating complex shapes like human poses (top) and overlapping instances (bottom). (c) Our holistic image boundary representation helps resolving ambiguities in a challenging scene.  }
\label{fig:Teaser}
\end{figure}


Deep learning has been widely used in various vision tasks~\cite{he2016deep,girshick2015fast,ren2015faster,zhang2019revisiting,long2015fully,zhang2020deepptz,kim2020video,zhang2020resnet,benz2020revisiting}. Most deep learning based instance segmentation methods follow the \textit{detect-then-segment} paradigm.
It first detects bounding boxes and then segments the instance mask in each bounding box.
However, this dominant paradigm has two fundamental issues.
First, it is step-wise, heavily relying on the accurate bounding box proposals.
Second, it produces low-resolution and coarse masks, resulting from only exploiting the region-specific information, \ie, RoI-pooled features.
To remedy these issues, recent studies attempt to incorporate a global representation called `basis' into the framework.
They consider combining the global image-level information, \ie, basis, with the original region-specific information in a single-stage and fully convolutional manner.


Despite their great effectiveness and flexibility, we argue that \textit{boundary} is a key representation for segmentation that is still missing in this basis framework.
Indeed, image boundaries are an important cue for recognition; Humans can recognize objects from sketches alone as instances in natural images are bounded by the edges. 
In the existing single-stage methods, the global mask representations are not explicitly learned in a supervised way, but loosely trained by a signal from the last 
mask output. However, such implicit learning of mask representations often ignores object shape and boundary information. This can lead to coarse and indistinct segmentation results because the pixels near boundaries are hard to be classified, as shown in \figref{fig:Teaser}. 
We are not the first one to exploit boundaries for instance segmentation. 
Recently, Boundary-preserving Mask RCNN \cite{ChengWHL20_boundary_mrcnn} proposed a boundary prediction head alongside the mask head of the two-stage Mask RCNN.
They have shown that the boundary representation can provide a more precise object location and mitigates its coarse and low-resolution predictions.
Motivated by this recent observation, we extend the idea in the context of the recent single-stage instance segmentation.
First, while the RoI-wise head of Boundary-preserving \cite{ChengWHL20_boundary_mrcnn} learns the boundary information for a single object at a time, we propose to learn holistic image-level instance boundaries.
Intuitively, heavy occlusions and shape complexities are better revealed at a whole image extent. 
We thus argue that learning the global boundary has its distinct advantages in segmenting individual objects in a scene, and will have a great synergy with the global mask representations in modern single-stage segmenters.
We name our proposed method as Boundary Basis based Instance Segmentation(\textbf{B2Inst}), and this new boundary basis is explicitly trained with the boundary ground-truths, which can be automatically generated from the given mask annotations. All masks and boundary representations are then cropped and composited together to form the final instance masks.
Second, instead of using the instance boundary information for representation learning only during the training, we use the model to learn a novel instance-wise boundary-aware mask score, which can be mainly utilized in test-time to prioritize high-quality masks. It is a generalized version of \cite{huang2019_MaskScore} and considers the agreement between the predicted masks and boundaries as well. It is a unified measure of both mask and boundary quality.


To sum up, we explicitly use image and instance boundaries as an additional important learning cue.
We find that both the proposed holistic image boundary representation and the boundary-aware mask score operate complementarily to each other, providing better localization performance and help to resolve ambiguities between overlapping objects, as shown in \figref{fig:Teaser}.  
We apply our \textit{B2Inst} to the strongest single-stage instance segmentation methods, \ie, BlendMask \cite{chen2020blendmask} and YOLACT \cite{bolya2019yolact}, and achieve consistent improvements on the challenging COCO benchmark.

The main contributions of this paper are as follows.

\begin{enumerate}[topsep=0pt,itemsep=0pt]
\item This is the first time to explore the boundary representation for accurate basis-based instance segmentation.
\item We propose \textit{B2Inst} which incorporates the boundary representation both on the global and the local view. 
At an image-level, we propose to learn holistic boundaries to enhance global basis representation. 
In an instance-level, we design a boundary-aware mask score. It captures both the mask and boundary quality at the same time and can prioritize high-quality mask predictions at test time. We show that both are complementary to each other in building an accurate segmenter.
\item We conduct extensive ablation studies to verify the effectiveness of our proposals. 
To further show that our algorithm is general, we apply it to the two strong basis-based instance segmentation frameworks and demonstrate consistent improvements both visually and numerically.
Finally, by augmenting the BlendMask framework with our method, we achieve new state-of-the-art results.
\end{enumerate}

\section{Related Work}

\subsection{Instance segmentation}
Most existing instance segmentation algorithms can be categorized into either two-stage or one-stage approaches. The two-stage methods first detect the bounding box for each object instance and then predict the segmentation mask inside the cropped box region. Among many previous studies \cite{hariharan2014simultaneous, dai2016_mnc, kirillov2017instancecut, li2017_fcis, chen2018_masklab, pinheiro2016learning, kim2020vps}, Mask RCNN \cite{He_2017_ICCV_MaskRcnn} is the most representative of this category, and it employs a proposal generation network and RoIAlign feature pooling strategies to obtain fixed-sized features of each proposal. Further improvements have been made to boost its accuracy; PANet \cite{liu2018path} introduces bottom-up path augmentation to enrich FPN features, and Mask scoring RCNN \cite{huang2019_MaskScore} addresses the misalignment between the confidence score and localization accuracy of predicted masks. These two-stage approach has dominated in state-of-the-art performances thus far in instance segmentation. However, these methods require RoI-wise feature pooling and head operations which make their inference quite slow, limiting the use in real-time applications.

Single-stage approaches typically aim at faster inference speed by avoiding proposal generation and feature pooling operations. In particular, bottom-up methods view the instance segmentation as a label-then-cluster problem. Instead of assigning RoI proposals, it produces pixel-wise predictions of cues such as directional vectors \cite{liu2017sgn}, pairwise affinity \cite{liu2018affinity},  watershed energy \cite{bai2017deep}, and embedding learning, and then group object instances from the cues in the post-processing stage. There are local-area-based methods that output instance masks on each local region directly \cite{chen2019tensormask, xie2020polarmask} without dependency on box detection or grouping post-processing. However, the single-stage approaches thus far have been lagging behind in accuracy compared to the two-stage counterparts, especially on the challenging dataset like COCO. The latest global-area-based methods \cite{chen2020blendmask,bolya2019yolact, Wang_centermask_2020_CVPR}  have broken the record, often outperforming the Mask RCNN \cite{He_2017_ICCV_MaskRcnn}. They first generate intermediate FCN feature maps, called `basis', then assemble the extracted basis features to form the final masks for each instance. 

In this work, we especially focus on the latest category of basis-based methods. More detailed review will be provided in \sref{sec:basis}.

\subsection{Boundary learning for instance segmentation}
Boundary detection has been a fundamental computer vision task as it provides an important cue for recognition \cite{chen2016semantic, hayder2017boundary, yuan2020segfix, cheng2020learning, yu2018learning, takikawa2019gated, huang2016object, bertasius2016semantic}. Accurate boundary localization is able to explicitly contribute to the mask prediction for segmentation. CASENet \cite{yu2017casenet} presents a challenging task of category-aware boundary detection. InstanceCut \cite{kirillov2017instancecut} adopts boundaries to partition semantic segmentation into instance-level segmentation. However, it involves expensive super-pixel extractions and grouping computations. Zimmermann et al. propose boundary agreement head \cite{zimmermann2019faster} to focus on instance boundaries with an auxiliary edge loss. Boundary preserving Mask RCNN \cite{ChengWHL20_boundary_mrcnn} proposes to predict instance-level boundaries to augment the mask head. These two-stage methods focus on the RoI-wise boundary of a single object, and thus lack a holistic view of image boundaries which can often resolve ambiguities in multiple overlapping objects and complex scenes. 



\begin{figure*}
\begin{tabular}{@{}c@{}}
\includegraphics[width=1.0\linewidth]{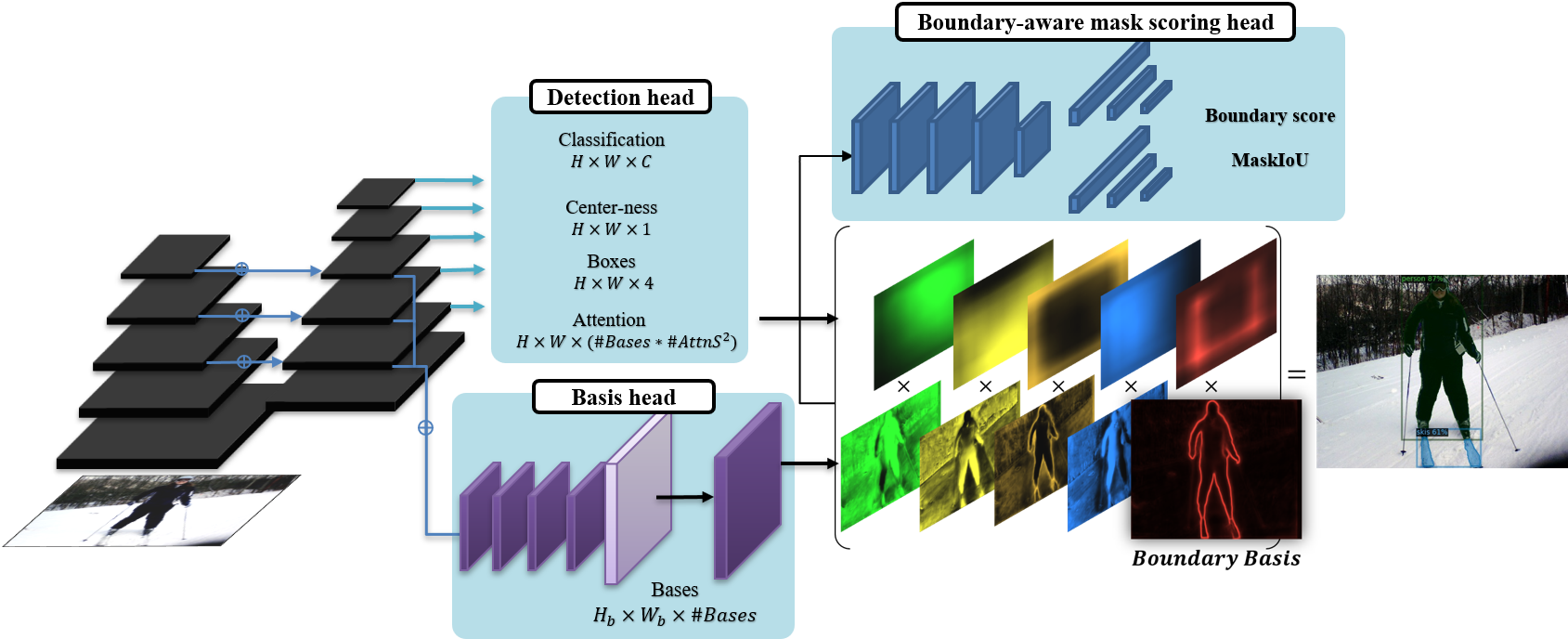}\\
\end{tabular}
\vspace{3mm}
\caption{ {\bf Overview of our method.} The figure shows the BlendMask \cite{chen2020blendmask} instantiation. 
It generates a set of basis mask representations. In parallel, it predicts instance-level parameters given the detected boxes, \textit{i.e.}, attention maps, which are used in combining the basis features into the final per-instance masks.
On top of this strong basis-based instance segmentation framework, we incorporate the boundary representation both on the global and the local view. 
At an image-level, the model learns to produce a holistic image boundary, improving the global basis features.
At an instance-level, the model predicts a boundary-aware mask score, which captures both the mask and boundary quality jointly and can prioritize high-quality mask predictions at test time.}
\label{fig:networks}
\end{figure*}


\section{Background}
\label{sec:basis}

\noindent\textbf{Basis-based instance segmentation:} 
There has been a wave of studies on single-stage instance segmentation, enabled by recent advances in one-stage detection such as FCOS, which outperforms the two-stage counterpart in accuracy. These dense instance segmentation methods generate a set of basis mask representations, called \textit{basis}, by a fully convolutional network. In parallel, there is another task of predicting instance-level parameters given the detected boxes, that are used in combining the basis features into the final per-instance masks. Recent basis-based instance segmentation methods follow this general pipeline \cite{bolya2019yolact, Cao_SipMask_ECCV_2020, Wang_centermask_2020_CVPR, chen2020blendmask, tian2020conditional}. 

First, YOLACT \cite{bolya2019yolact} presents to predict 32 global bases and the according instance-specific \textit{scalar} coefficients. The final mask is then computed by a linear combination among the bases. 
After YOLACT, the community has pushed either in the direction of developing rich instance features or designing effective assembly methods.
BlendMask \cite{chen2020blendmask}, SipMask \cite{Cao_SipMask_ECCV_2020}, and CenterMask \cite{Wang_centermask_2020_CVPR} improved the assembly parameters, by extending from the scalar to \textit{spatial} 2D coefficient matrices.
CondInst \cite{tian2020conditional} considers the \textit{dynamic filter weights} as implicit instance-specific representation. The mask is then predicted by a standard sliding window operation over the basis features, with the predicted convolution weight filters.

In this work, we take one step further by investigating the under-explored boundary information for both the basis and instance-specific features. 
We first propose to learn holistic boundaries to enhance the global basis representation. 
Then, we design a boundary-aware mask score to unify and jointly learn the mask and boundary quality of an object instance.
Putting together, our method consistently improves over the latest basis-based instance segmentation methods, which implies the generality and orthogonality of our methods to the existing ones.

\section{Exploiting Boundary Representation}
Overall architecture of our \textit{B2Inst} method is shown in \figref{fig:networks}. 
It consists of four parts: $(1)$ backbone for feature extraction, $(2)$ instance-specific detection head, $(3)$ global basis head, and $(4)$ mask scoring head.
The final masks are computed by combining the instance-specific information from the detection head and the global image information from the basis head.
We note that \figref{fig:networks} shows the BlendMask instantiation.


\subsection{Learning holistic image boundary}

\noindent\textbf{Standard basis head}:
The basis head takes FPN \cite{Lin_FPN_2017_CVPR} features as input and produces a set of $K$ global basis features. It consists of four consecutive $3\times 3$ convolution layers followed by an upsampling and the last layer that reduces the channel-size to $ K $. The previous basis head \cite{bolya2019yolact,Cao_SipMask_ECCV_2020, chen2020blendmask,tian2020conditional} is supervised by the last mask loss only, and thus the basis representations are obtained in an unsupervised manner.

\noindent\textbf{Missing piece - image boundary}:
There is a notable recent study in two-stage framework that improves instance segmentation by learning RoI-wise boundary information. Different from them, we propose to learn a holistic boundary of all instances in a scene, instead of a single instance one by one. Overlapping objects and complex shapes are more clearly captured given the whole image context (see \figref{fig:Teaser}). Moreover, the boundary supervision comes at no cost from the given mask annotations. 


\noindent\textbf{Boundary ground-truths}:
We use the Laplacian operator to generate soft boundaries from the binary mask ground-truths.  The boundaries are then thresholded at 0 and converted into final binary maps.

\noindent\textbf{Objective function}:
We use three different loss functions in total: 1) binary cross-entropy loss, 2) dice loss, and 3) boundary loss \cite{deng2018_boundary_loss}. Especially, the boundary loss is proposed to predict sharp boundaries at test time without any sophisticated post-processing.


\subsection{Boundary-aware mask scoring}
Mask scoring R-CNN \cite{huang2019_MaskScore} proposes a mask IoU scoring module instead of using the classification score to evaluate the mask, which can improve the quality of mask results. The mask IoU (Jaccard index) evaluates the agreement between the mask prediction and ground-truth by focusing only on their \textit{area}, without considering their \textit{shapes}. The discrepancy between the mask's IoU score and visual quality can be found in the top-performing segmentation method~\cite{chen2020blendmask} (\figref{fig:comparisonOFQualitative}-a). Between the mask predictions and ground-truths, their \textit{areas} agree on most pixels, while their \textit{boundaries} do not. Such a problem becomes even more severe when there are overlapping objects and fine details (row 1 and 3), yet is not able to be quantified by the current mask IoU metric. To address this issue, we devise a mask score that can respect both the IoU and boundary quality.

        
        
    

\noindent\textbf{Boundary score}: 
We define $S_{boundary}$ to evaluate the boundary agreement between mask prediction $M_{pred}$ and ground-truth $M_{gt}$. We use a Dice metric~\cite{deng2018_boundary_loss} which is insensitive to the number of foreground and background pixels, and thus can have balanced importance on the boundary pixels. The Dice metric also has a range from 0 to 1, which makes it a comparable scoring function to the IoU metric. To compute $S_{boundary}$, we first identify the boundaries in the predicted and ground-truth masks by filtering them with a Laplacian kernel $\Delta f$ as $B_{pred} = \Delta f(M_{pred})$ and $B_{gt} = \Delta f(M_{gt})$. Then, the Dice metric is computed between the two mask boundaries as in \eref{eq:DiceCoeff}.

\begin{equation} \label{eq:DiceCoeff}
S_{boundary} =\frac{2\sum_{i}^{h \times w} B_{pred}^{i}B_{gt}^{i}+\epsilon}{\sum_{i}^{h \times w} (B_{pred}^{i})^2 + \sum_{i}^{h \times w}(B_{gt}^{i})^2+\epsilon}
\end{equation}

where $i$ stands for the $i$-pixel and $\epsilon$ is a soft-term to avoid division of zero, and $h$ and $w$ are the height and width of the predicted bounding box size.




$S_{mask}$ can be divided into 1) a classification problem matching the correct class and 2) a problem of regressing the maskIoU in the proposal of the corresponding class. 
In the first task, $S_{class}$ is the confidence score of which class the bounding box proposal corresponds to. 
$S_{class}$ is in charge of the classification task at the proposal stage. 
The other tasks, $S_{IoU}$ is regressed using an additional header in MS-RCNN \cite{huang2019_MaskScore} 
The object mask and its boundary are complementary, so we can easily change either one to another. 
Therefore, we designed a simple network that regresses $S_{IoU}$ and $S_{Dice}$ at the same time. 
This part will be covered in the next paragraph.

\noindent\textbf{Scoring head}:
We introduce a unified mask scoring head that can predict the mask scores during training. As we want to incorporate the boundary score $S_{boundary}$ with the conventional mask IoU score $S_{IoU}$, our score prediction head is trained to regress the two scores at a time. To fully exploit the correlation between mask boundary and IoU information, it is designed to be a shared header of 4 convolution layers with last fully-connected branches for each $S_{IoU}$ and $S_{boundary}$ regression tasks. The inputs of our scoring head are a concatenation of predicted mask $M_{pred}$, boundary $B_{pred}$, and RoI-pooled FPN features $F_{RoI}$, all in the shape of $\mathbb{R}^{h \times w}$. We use the ground-truth bounding boxes during the training of our boundary-aware mask scoring head.





\noindent\textbf{Score definition at inference}:
At inference, we define the mask score $S_{mask}$ using $S_{IoU}$ and $S_{boundary}$ that are predicted from our score head. We follow the standard inference procedure~\cite{He_2017_ICCV_MaskRcnn}; First, the top-k scoring proposals from the detection head are fed into our scoring head, then the mask scores are produced. Finally, our mask confidence score is computed by multiplying the $S_{IoU}$ and $S_{boundary}$ with the standard classification score $S_{mask}$ of the bounding box as follows:

\begin{equation} \label{eq:BMscoring}
S_{mask} =  S_{class} \cdot {\scriptstyle\sqrt{S_{IoU} \cdot S_{boundary}}}.
\end{equation}

\section{Experiment}
In this section, we conduct experiments to analyze our two major proposals in \textit{B2Inst} : 1) holistic boundary basis
and 2) boundary-aware mask scoring.
Specifically, we perform ablation studies in~\sref{sec:abl}.
We then incorporate our proposals into the two strong basis-based instance segmentation frameworks in~\sref{sec:other_basis}.
Finally, we evaluate our mask results quantitatively and qualitatively, compared with the state-of-the-art baselines in~\sref{sec:main_exp}.

\noindent\textbf{Dataset \& Evaluation Metric}
We present experimental results on the MS COCO instance segmentation benchmark \cite{lin2014microsoft}. 
All the experiments conducted here follow standard COCO 2017 setup, using 115k images as $train$ split, 5k images as $validation$ split. 
We conduct detailed ablation experiments on the validation split using BlendMask~\cite{chen2020blendmask} framework.
We then report the final COCO mask AP on 20k $test$-$dev$ split. We use the official evaluation server as there are no labels released to the public.
For the evaluation metric, we use standard COCO-style AP. The results include AP at different scales, $AP_{S}$, $AP_{M}$, $AP_{L}$.


\subsection{Ablation study}
\label{sec:abl}

\subsubsection{Impact of two major proposals}
The main components of our framework design are the holistic boundary basis and boundary-aware mask-scoring.
The ablation results are summarized in~\tabref{tab:abl_module}.

\noindent\textbf{Holistic Boundary basis (HBB)}
To see the holistic boundary feature's impact, we add it to the original global basis representations.
The performance of \textit{B2Inst} with additional boundary information improves by $0.4$ $AP$ overall.
The improvement of $AP_{75}$ and $AP_{S}$ are relatively big, implying that the boundary information effectively guides the model to predict more precise masks in challenging cases such as precise localization or small-object segmentation.
Note that the cost of obtaining the ground truth boundary is negligible.


\noindent\textbf{Boundary-aware mask scoring (BS~+~MS)}
We generalize the original mask score~\cite{huang2019_MaskScore} to consider the agreement between the predicted masks and boundaries. In this way, the scoring metric can capture both the mask (\textit{area}) and boundary (\textit{shape}) quality jointly.
As shown in the table, both scoring metrics can push the performance, demonstrating their efficacy.
Further improvement comes when using both together (by $0.6$ $AP$), meaning that both the mask and boundary quality are crucial when evaluating the mask quality at test time.


In a brief conclusion, we show that the holistic boundary basis improves global basis representation. It especially resolves the ambiguities in challenging cases. Moreover, the proposed boundary-aware mask scoring well prioritizes high-quality masks at test time, demonstrating the efficacy of considering both the area and shape quality jointly.
We show that both proposals operate complementarily to each other, contributing to the final performance improved consistently. 

\begin{table}[t]
\begin{center}
    \resizebox{0.47 \textwidth}{!}{
    \begin{tabular}{ccc|cccccc}
    \hline\hline
    \multicolumn{1}{c}{\textbf{HBB}} & \multicolumn{1}{c}{\textbf{BS}} & \multicolumn{1}{c|}{\textbf{MS}} & \textbf{AP}   & \textbf{AP50} & \textbf{AP75} & \textbf{APs} & \textbf{APm} & \textbf{APl} \\
    \hline
                                    &                                 &                                  & \textbf{35.8} & 56.4          & 38.1          & 16.9         & 38.8         & 51.1         \\
    \checkmark                      &                                 &                                  & \textbf{36.2} & 56.5          & 38.5          & 17.6         & 39.3         & 51.3         \\
    \checkmark                      & \checkmark                      &                                  & \textbf{36.3} & 55.3          & 39.4          & 16.8         & 39.3         & 52.3         \\
    \checkmark                      &                                 & \checkmark                       & \textbf{36.6} & 55.5          & 39.6          & 17.3         & 39.7         & 51.9         \\
    \checkmark                      & \checkmark                      & \checkmark                       & \textbf{36.8} & 55.7          & 39.9          & 17.2         & 39.9         & 52.5 
    \end{tabular}
}
\end{center}
\vspace{1mm}
   \caption{\textbf{Impact of two major proposals.} 
   We verify the efficacy of two major proposals, holistic boundary basis, and boundary-aware mask scoring.
   HBB, BS, and MS denote Holistic Boundary Basis, Boundary Dice Coefficient Scoring, and Mask IoU Scoring, respectively. We observe that both proposals contribute to the final performance advanced.}
\label{tab:abl_module}
\vspace{-10pt}
\end{table}

\begin{table}[h]
\begin{center}
    \resizebox{0.46 \textwidth}{!}{
    \begin{tabular}{c|cccccc}
    \hline\hline
    \multicolumn{1}{c|}{\textbf{Design}} & \textbf{AP}   & \textbf{AP50} & \textbf{AP75} & \textbf{APs} & \textbf{APm} & \textbf{APl} \\ \hline
    \textbf{Baseline}                    & 35.8          & 56.4          & 38.1          & 16.9         & 38.8         & 51.1         \\
    \textbf{(a) }          & 35.7          & 56.2          & 38.0          & 17.3         & 38.6         & 50.9         \\
    \textbf{(b) }             & 36.0          & 56.4          & 38.4          & 16.4         & 39.0         & 51.8         \\
    \textbf{(c) }               & 36.0          & 56.5          & 38.5          & 16.5         & 39.0         & 51.6         \\
    \textbf{(d) }                    & 36.0          & 56.2          & 38.4          & 16.7         & 38.7         & 51.5         \\
    \textbf{(e) }          & \textbf{36.2} & 56.5          & 38.5          & 17.6         & 39.3         & 51.3   
    \end{tabular}}
\end{center}
\vspace{1mm}
   \caption{\textbf{Basis head design choices.} 
   We compare five different ways of configuring the basis head to predict boundary representation.
   The details of each configurations are elaborated in~\sref{sec:basis_head}.
   The experiments are conducted on COCO $val2017$.}
\label{tab:design_choice_BB}
\vspace{-10pt}
\end{table}

\begin{figure}[t]
\small
\begin{tabular}{@{}c}
\includegraphics[width=1\linewidth]{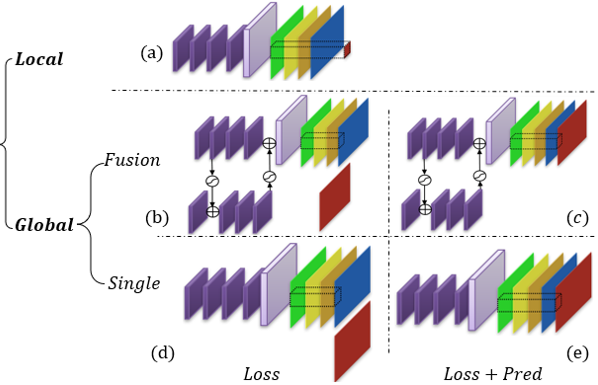}
\end{tabular}
\vspace{1mm}
\caption{{\bf Five different ways to design the basis head.}
    \textit{Local} and \textit{Global} indicate whether we use boundary supervision locally or globally.
    \textit{Fusion} and \textit{Single} indicate whether we use multi-scale feature fusion~\cite{ChengWHL20_boundary_mrcnn} or not.
    \textit{Loss~+~Pred} denotes that we use both the boundary supervision and its prediction.
    Green, Yellow, Orange, Blue, and Red images represent the original basis. 
    The Red image represents the additional image boundary basis.}
\label{fig:design_choice_BB}
\vspace{-10pt}
\end{figure}

\begin{table*}[t]
\begin{center}
    \resizebox{0.9 \textwidth}{!}{

\begin{tabular}{l|c|c|ccccc|c|ccccc}
\hline\hline
\multicolumn{1}{c|}{\multirow{2}{*}{\textbf{Design}}} & \multirow{2}{*}{\textbf{FPS}} & \multicolumn{6}{c|}{\textbf{Mask}}                                                          & \multicolumn{6}{c}{\textbf{BBOX}}                                                        \\ \cline{3-14} 
                                 &                               & \textbf{AP}   & \textbf{AP50} & \textbf{AP75} & \textbf{APs} & \textbf{APm} & \textbf{APl}  & \textbf{AP} & \textbf{AP50} & \textbf{AP75} & \textbf{APs} & \textbf{APm} & \textbf{APl}  \\ \hline
YOLACT                  & 45.3                 & 28.1          & 46.2          & 29.0          & 9.7          & 30.4         & 46.4          & 30.2                 & 50.6          & 31.6          & 14.5         & 32.3         & 44.2 \\
\textbf{B2Inst}-YOLACT           & 42.5                 & \textbf{30.1} & 48.8          & 31.6          & 10.7         & 32.2         & 48.1          & \textbf{31.1}        & 53.0          & 32.3          & 14.6         & 34.5         & 44.8 \\

\hline

BlendMask*              & 10.5                 & 35.8          & 56.4          & 38.1          & 16.9         & 38.8         & 51.1          & 40.2                 & 59.2          & 43.6          & 24.2         & 43.8         & 51.5 \\ 
\textbf{B2Inst}-BlendMask*       & 9.7                  & \textbf{36.8} & 55.7          & 39.9          & 17.2         & 39.9         & 52.5          & \textbf{40.2}        & 59.0          & 43.7          & 23.5         & 43.6         & 52.3         
\end{tabular}

}
\end{center}
\vspace{0mm}
  \caption{\textbf{Combining with existing basis-based methods.} In order to show that our approach is general, we apply our proposals to the two different state-of-the-art basis-based methods, YOLACT~\cite{bolya2019yolact} and BlendMask~\cite{chen2020blendmask}.
  The experiments are conducted on COCO $val2017$.
  We see that our method consistently improves the baseline performances clearly with only a slight difference in \textit{FPS}.
  }
\label{tab:abl_architecture}
\vspace{-12pt}
\end{table*}


\subsubsection{Basis head design choices}
\label{sec:basis_head}
In this experiment, we compare five different ways of designing the basis head to predict boundary representation (see \figref{fig:design_choice_BB}). The results are shown in \tabref{tab:design_choice_BB}.
We describe each configuration in the following:
\begin{itemize}[topsep=1pt,itemsep=1pt]
\item \textbf{(a)} The boundary loss is applied locally, \textit{i.e.}, on the RoI pooled basis feature.
\item \textbf{(b)} Following Boundary Mask-RCNN \cite{ChengWHL20_boundary_mrcnn}, multi-scale feature fusion is adopted. In addition, the boundary loss is applied globally, \textit{i.e.}, on the whole basis feature.
\item \textbf{(c)} (b)~+~The boundary prediction is explicitly used for the final per-instance mask composition.
\item \textbf{(d)} No multi-scale feature fusion is adopted. The boundary loss is applied globally.
\item \textbf{(e)} (d)~+~The boundary prediction is explicitly used for the final per-instance mask composition.

\end{itemize}

We observe that the global boundary supervision is important and also utilizing the boundary prediction as an additional basis representation is effective. We do not see any improvement when using the multi-scale feature aggregation.
The configuration (e) produces the best performance, and we use this setup in the following experiments.

\subsection{Combining with existing basis-based methods}
\label{sec:other_basis}
We see our proposals are general, thus can be easily applied to the existing basis-based methods.
In this experiment, we combine our two major proposals with two different basis-based approaches: YOLACT~\cite{bolya2019yolact} and BlendMask~\cite{chen2020blendmask}. The results are summarized in~\tabref{tab:abl_architecture}.
We observe that our method consistently improves the performance of all the baselines.
More specifically, our proposals improve those frameworks by $2.0$ $AP$ and $1.0$ $AP$, respectively.
The positive results imply that the boundary representation is indeed a fundamental representation for recognition and is missing in the previous basis-based frameworks.
We show that the proposed concept of both the holistic boundary representation and the boundary-aware mask score effectively addresses the issue.

\subsection{Main results}
\label{sec:main_exp}
\subsubsection{Quantitative results}
We compare our method with several state-of-the-art instance segmentation methods, including both the two-stage \cite{dai2016_mnc, li2017_fcis,chen2018_masklab, He_2017_ICCV_MaskRcnn, huang2019_MaskScore, ChengWHL20_boundary_mrcnn} and one-stage \cite{chen2019tensormask, tian2020conditional,kirillov2020pointrend, chen2020blendmask,bolya2019yolact}.
All methods are trained on COCO $train2017$ and evaluated on COCO $test$-$dev2017$.
\tabref{tab:main_result} shows the results.
Without bells and whistles, we can surpass the existing state-of-the-art methods with the same ResNet-50 and ResNet-101 backbones.
In particular, Mask R-CNN \cite{He_2017_ICCV_MaskRcnn} and MS R-CNN \cite{huang2019_MaskScore} achieve overall mask AP scores of 34.6 and 35.8 in R-50 backbone, respectively. 
The recently introduced BMask R-CNN \cite{ChengWHL20_boundary_mrcnn} obtain mask AP scores of 35.9. 
Our method achieves 36.9 on the same R-50 backbone.

In case of single-stage methods, YOLACT-700 \cite{bolya2019yolact} and TensorMask \cite{chen2019tensormask} obtain a mask AP of 31.2 and 37.3 in R-101 backbone, respectively. 
CondInst \cite{tian2020conditional} and BlendMask* achieve a mask AP score of 39.1 and 39.5, respectively.  
CenterMask \cite{lee2019centermask} and PointRend \cite{kirillov2020pointrend} obtain the best results among the existing single-stage methods with a mask AP score of 39.8.
Under the same setting, \textit{i.e.}, input size and backbone, our method outperforms CenterMask \cite{lee2019centermask} and PointRend \cite{kirillov2020pointrend} with a healthy margin (+~1.0). 

\begin{table*}[t!]
\centering
\resizebox{0.7 \textwidth}{!}{
\begin{tabular}{c|l|c|c|llllll}
\hline\hline
\textbf{Backbone}                    & \multicolumn{1}{c|}{\textbf{Method}}        & \multicolumn{1}{c|}{\textbf{epochs}} & \multicolumn{1}{c|}{\textbf{Aug}} & \multicolumn{1}{c}{\textbf{AP}}   & \multicolumn{1}{c}{\textbf{AP50}} & \multicolumn{1}{c}{\textbf{AP75}} & \multicolumn{1}{c}{\textbf{APs}} & \multicolumn{1}{c}{\textbf{APm}} & \multicolumn{1}{c}{\textbf{APl}} \\
\hline
\multirow{12}{*}{\textbf{R-50\_FPN}} & \multicolumn{1}{c|}{\textbf{Two-Stage:}}    &                                      &                          &                                   &                                   &                                   &                                  &                                  &                                  \\
                                     & Mask R-CNN                                  & 12                                   &                         & 34.6                              & 56.5                              & 36.6                              & 15.4                             & 36.3                             & 49.7                             \\
                                     & MS R-CNN                                    & 12                                   &                         & 35.8                              & 56.5                              & 38.4                              & 16.2                             & 37.4                             & 51.0                             \\
                                     & BMask R-CNN                                 & 12                                   &                         & 35.9                              & 57.0                              & 38.6                              & 15.8                             & 37.6                             & 52.2                             \\\cline{2-10}
                                     & \multicolumn{1}{c|}{\textbf{Single-Stage:}} &                                      &                          &                                   &                                   &                                   &                                  &                                  &                                  \\
                                     & TensorMask                                  & 72                                   & \multicolumn{1}{c|}{\checkmark}   & 35.5                              & 57.3                              & 37.4                              & 16.6                             & 37.0                             & 49.1                             \\
                                     & CondInst                                    & 12                                   & \multicolumn{1}{c|}{\checkmark}   & 35.9                              & 56.4                              & 37.6                              & 18.4                             & 37.9                             & 46.9                             \\
                                     & PointRend                                   & 12                                   &                                  & 36.3                              & -                                 & -                                 & -                                & -                                & -                                \\
                                     & Blendmask                                   & 12                                   &                                  & 34.3                              & 55.4                              & 36.6                              & 14.9                             & 36.4                             & 48.9                             \\
                                     & Blendmask*                                  & 12                                   & \multicolumn{1}{c|}{\checkmark}   & 35.8                              & 56.4                              & 38.3                              & 16.9                             & 38.8                             & 51.1                             \\
                                     & \textbf{B2Inst}-Blend                               & 12                                   & \multicolumn{1}{c|}{\checkmark}   & \multicolumn{1}{c}{\textbf{36.9}} & 55.6                              & 40.0                              & 17.3                             & 39.9                             & 52.6                             \\
                                     & \textbf{B2Inst}-Blend                               & 36                                   & \multicolumn{1}{c|}{\checkmark}   & \multicolumn{1}{c}{\textbf{39.0}} & 58.7                              & 42.3                              & 19.9                             & 42.0                             & 55.4                             \\ 
\hline\hline
\multirow{19}{*}{\textbf{R-101\_FPN}}         & \multicolumn{1}{c|}{\textbf{Two-Stage:}}    &                                      &                                   &                                   &                                   &                                   &                                  &                                  &                                  \\
                                     & MNC                                         & 12                                   &                                  & 24.6                              & 44.3                              & 24.8                              & 4.7                              & 25.9                             & 43.6                             \\
                                     & FCIS                                        & 12                                   &                                  & 29.2                              & 49.5                              & -                                 & -                                & -                                & -                                \\
                                     & FCIS++                                      & 12                                   &                                  & 33.6                              & 54.5                              & -                                 & -                                & -                                & -                                \\
                                     & MaskLab                                     & 12                                   &                                  & 35.4                              & 57.4                              & 37.4                              & 16.9                             & 38.3                             & 49.2                             \\
                                     & Mask R-CNN                                  & 12                                   &                                  & 36.2                              & 58.6                              & 38.4                              & 16.4                             & 38.4                             & 52.1                             \\
                                     & MS R-CNN                                    & 12                                   &                                  & 37.5                              & 58.7                              & 40.2                              & 17.2                             & 39.5                             & 53.0                             \\
                                     & BMask R-CNN                                 & 12                                   &                                  & 37.7                              & 59.3                              & 40.6                              & 16.8                             & 39.9                             & 54.6                             \\
                                     & Mask R-CNN*                                 & 36                                   & \multicolumn{1}{c|}{\checkmark}   & 38.8                              & 60.9                              & 41.9                              & 21.8                             & 41.4                             & 50.5                             \\\cline{2-10}
                                     & \multicolumn{1}{c|}{\textbf{Single-Stage:}} &                                      &                                   &                                   &                                   &                                   &                                  &                                  &                                  \\
                                     & YOLACT-700                                  & 54                                   & \multicolumn{1}{c|}{\checkmark}   & 31.2                              & 50.6                              & 32.8                              & 12.1                             & 33.3                             & 47.1                             \\
                                     & TensorMask                                  & 72                                   & \multicolumn{1}{c|}{\checkmark}   & 37.3                              & 59.5                              & 39.5                              & 17.5                             & 39.3                             & 51.6                             \\
                                     & CondInst                                    & 36                                   & \multicolumn{1}{c|}{\checkmark}   & 39.1                              & 60.9                              & 42.0                              & 21.5                             & 41.7                             & 50.9                             \\
                                     & Centermask*                                 & 36                                   & \multicolumn{1}{c|}{\checkmark}   & 39.8                              & -                                 & -                                 & 21.7                             & 42.5                             & 52.0                             \\
                                     & PointRend                                   & 36                                   & \multicolumn{1}{c|}{\checkmark}   & 39.8                              & -                                 & -                                 & -                                & -                                & -                                \\
                                     & Blendmask                                   & 36                                   & \multicolumn{1}{c|}{\checkmark}   & 38.4                              & 60.7                              & 41.3                              & 18.2                             & 41.5                             & 53.3                             \\
                                     & Blendmask*                                  & 36                                   & \multicolumn{1}{c|}{\checkmark}   & 39.5                              & 61.1                              & 42.3                              & 19.5                             & 42.9                             & 56.2                             \\
                                     & \textbf{B2Inst}-Blend                               & 12                                   & \multicolumn{1}{c|}{\checkmark}   & \multicolumn{1}{c}{\textbf{38.9}} & 58.4                              & 42.1                              & 18.4                             & 42.3                             & 55.6                             \\
                                     & \textbf{B2Inst}-Blend                               & 36                                   & \multicolumn{1}{c|}{\checkmark}   & \multicolumn{1}{c}{\textbf{40.8}} & 61.0                              & 44.2                              & 19.8                             & 44.1                             & 58.3                            
\end{tabular}
}

\vspace{1.5mm}
   \caption{\textbf{Quantitative comparisons with the state-of-the-art methods.} We use MS-COCO $test$-$dev2017$ to obtain the final performances. We compare our method with several state-of-the-art instance segmentation methods, including both the two-stage and single-stage. * indicates the reproduced performance in our $Detectron2$ \cite{wu2019detectron2} platform. 'Aug.' denotes using multi-scale training with shorter side range in [640, 800].}
\label{tab:main_result}
\vspace{-12pt}
\end{table*}

\begin{figure*}
\begin{tabular}{@{}c@{}}
\includegraphics[width=1.0\linewidth]{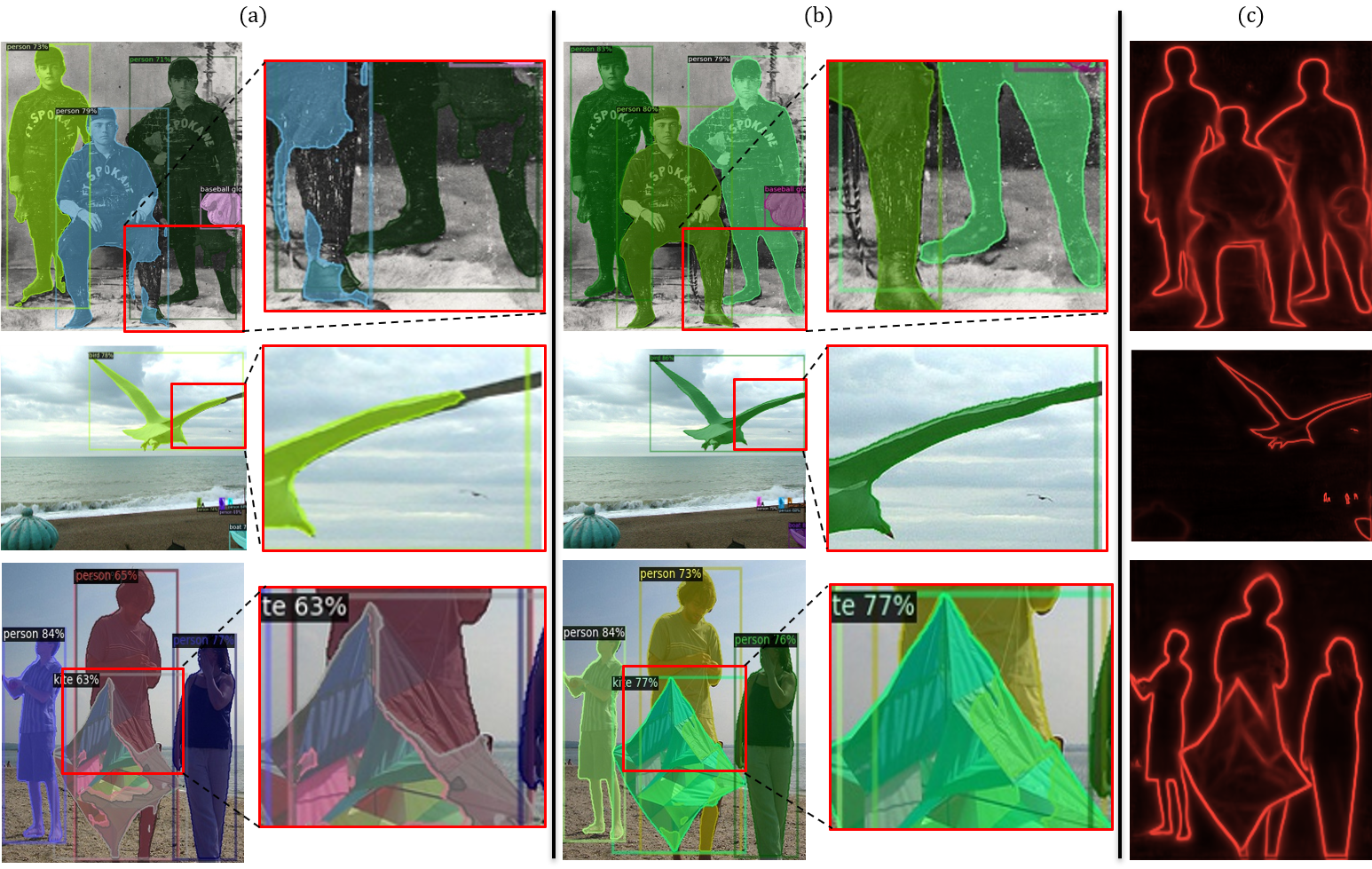}\\
\end{tabular}
    \caption[Optional caption for list of figures]{
    \textbf{Qualitative Comparisons on the Boundary Area.} We compare the mask results of (a) BlendMask \cite{chen2020blendmask} and (b) Our proposed \textit{B2Inst}. (c) Our learned image boundary representation.
    We can clearly see that our proposals resolve ambiguities in a challenging scene and produce more accurate masks.
    }
\label{fig:comparisonOFQualitative}
\end{figure*}

\subsubsection{Qualitative results}
\noindent\textbf{Improved Boundary Prediction}
\figref{fig:comparisonOFQualitative} shows visual mask results of BlendMask \cite{chen2020blendmask} and ours on the COCO validation set.
The first column (a) is the BlendMask result, and the second column (b) is our result. The third column (c) is the corresponding holistic image boundary bases that are used in (b) to predict final masks. 
The first image is monochrome photography with the salt and pepper noise.
BlendMask is vulnerable to this noise, and thus the object and its surroundings are not delineated accurately.
In the second image, BlendMask misses predicting the gull's wing on the boundary area.
In the third image, BlendMask fails to predict the kite area correctly due to the overlapping similar textures.
On the other hand, we can clearly see that our result has a much better mask prediction quality, especially in the boundary area. 

\begin{figure*}
\begin{tabular}{@{}c@{}}
\includegraphics[width=1\linewidth]{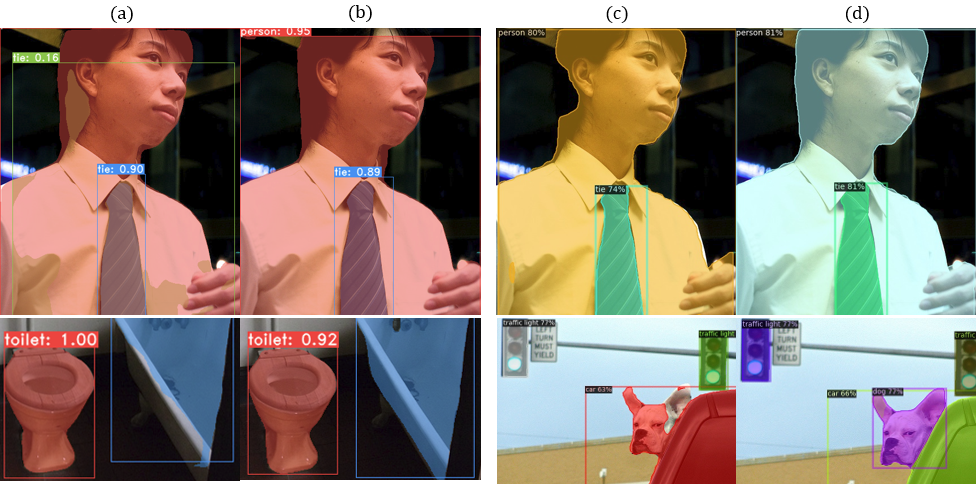}\\
\end{tabular}
\vspace{1mm}
    \caption[]{\textbf{Qualitative Comparisons between the Baselines \cite{bolya2019yolact,chen2020blendmask} and Ours on the COCO validation set.} (a) is the YOLACT result and (b) is the result of \textit{B2Inst}-YOLACT. Similarly, (c) is the BlendMask result, and (d) is the result of \textit{B2Inst}-BlendMask.}
\label{fig:compYOLACT_BLEND}
\end{figure*}

\noindent\textbf{Improved Mask Prediction}
\figref{fig:compYOLACT_BLEND} compares the  visual mask results of YOLACT \cite{bolya2019yolact}, BlendMask \cite{chen2020blendmask} and ours.
The first column (a) is the YOLACT result, and the second column (b) is the result of \textit{B2Inst}-YOLACT. 
Similarly, the third column (c) is the BlendMask result, and the fourth column (d) is the result of \textit{B2Inst}-BlendMask.
Again, we observe that our proposals successfully guide the baseline models to better capture the target object.




\section{Conclusion}
In this paper, we propose \textit{B2Inst} which learns an image boundary representation to enhance instance segmentation. 
We build upon the single-stage framework which is drawing much attention nowadays due to its speed and higher accuracy. 
Our \textit{B2Inst} learns a boundary basis representation by an explicit boundary prediction learning. 
In particular, the boundary basis is assembled together with the existing mask basis representations to compute the final per-instance masks, providing a better understanding of occlusions and complex shapes in a scene. In addition, we present a quality metric that evaluates both the mask and boundary prediction of each object and trains our model to predict the unified score that can be used in mask ranking at test time. Benefiting from these two proposed components, our model delineates instance shapes more accurately, especially for heavily occluded and complex scenes. When applied to BlendMask and YOLACT, which are the strongest single-stage methods, we exhibit consistent improvements over them both visually and numerically. Our model outperforms the state-of-the-art segmenters both for single-stage and two-stage frameworks. We hope our initial findings of the importance of boundary representation invigorate the follow-up studies.

\section*{Acknowledgment}
This work was supported by the Technology Innovation Program: Development of core technology for advanced locomotion/manipulation based on high-speed/power robot platform and robot intelligence (No.10070171) funded By the Ministry of Trade, industry \& Energy(MI, Korea)

\clearpage
{\small
\bibliographystyle{ieee_fullname}
\bibliography{egbib}
}

\section{Implementation details}
We verify our method mainly on the BlendMask \cite{chen2020blendmask} framework.
Most hyperparameters are kept the same except the resolution of attention map $7 \times 7$ for the training efficiency.
ResNet-50 \cite{he2016deep} is used as our backbone network and the weights are pre-trained on ImageNet \cite{krizhevsky2012imagenet}.
We train the model using stochastic gradient descent (SGD) with 4 TITAN RTX GPUs. 
We set the mini-batch size to 16 images.
We adopt $1\times$ schedule (90K iterations) with the initial learning rate of 0.01.
It is reduced by a factor of 10 at iteration 60$K$ and 80$K$, respectively.
Input images are resized to have shorter side 800 pixels and a longer side at maximum 1333 pixels.

\section{Qualitative Results}

\begin{figure*}
\begin{tabular}{@{}c@{}}
\includegraphics[width=1\linewidth]{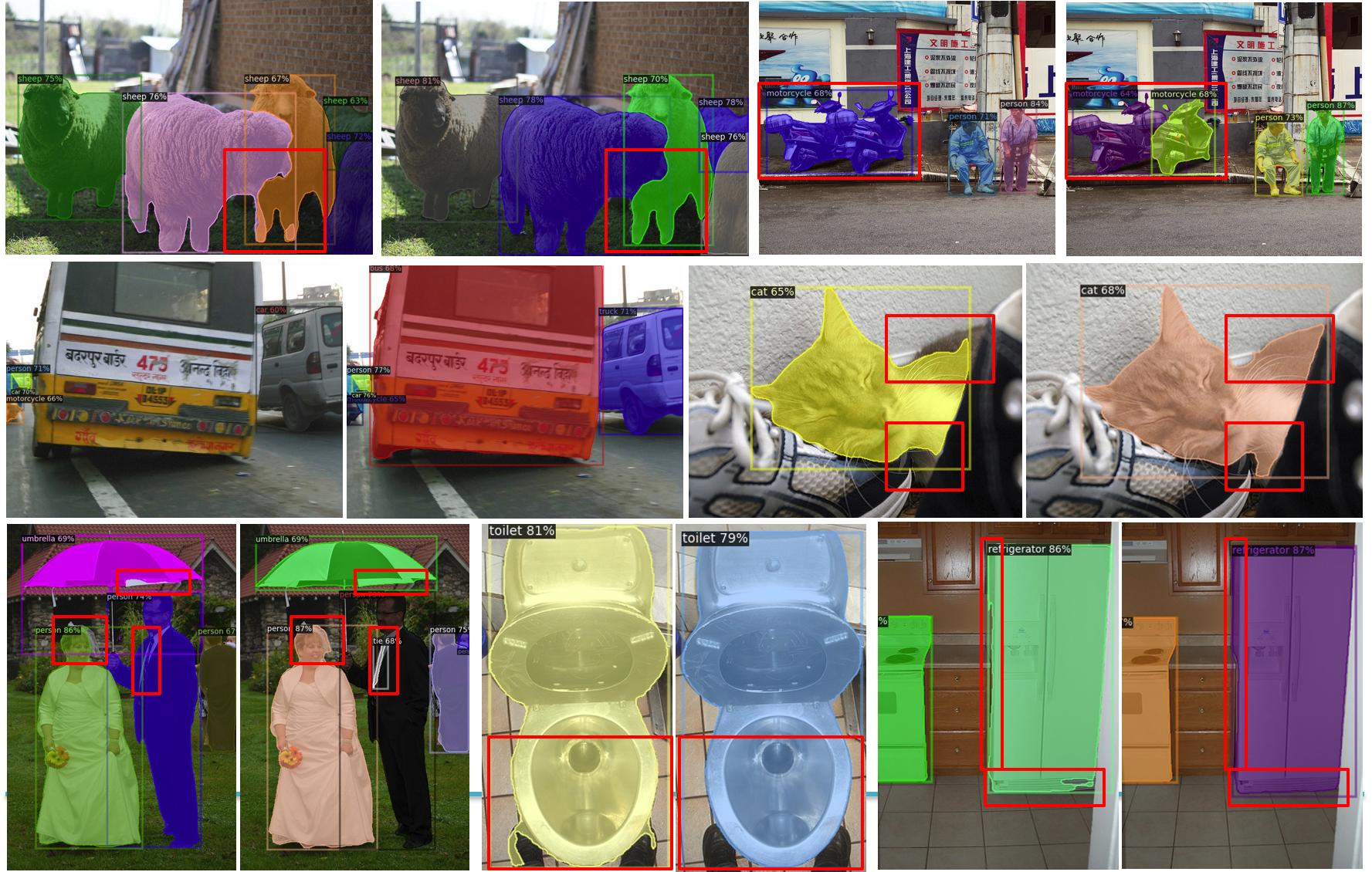}\\
\end{tabular}
\vspace{1mm}
    \caption[]{\textbf{Qualitative Comparisons between BlendMask \cite{chen2020blendmask} and \textit{B2Inst}-BlendMask.} 
    The image on the left is from the BlendMask while the image on the right is from ours.}
\label{fig:sup_blend2ours}
\vspace{-5pt}
\end{figure*}

\begin{figure*}
\begin{tabular}{@{}c@{}}
\includegraphics[width=1\linewidth]{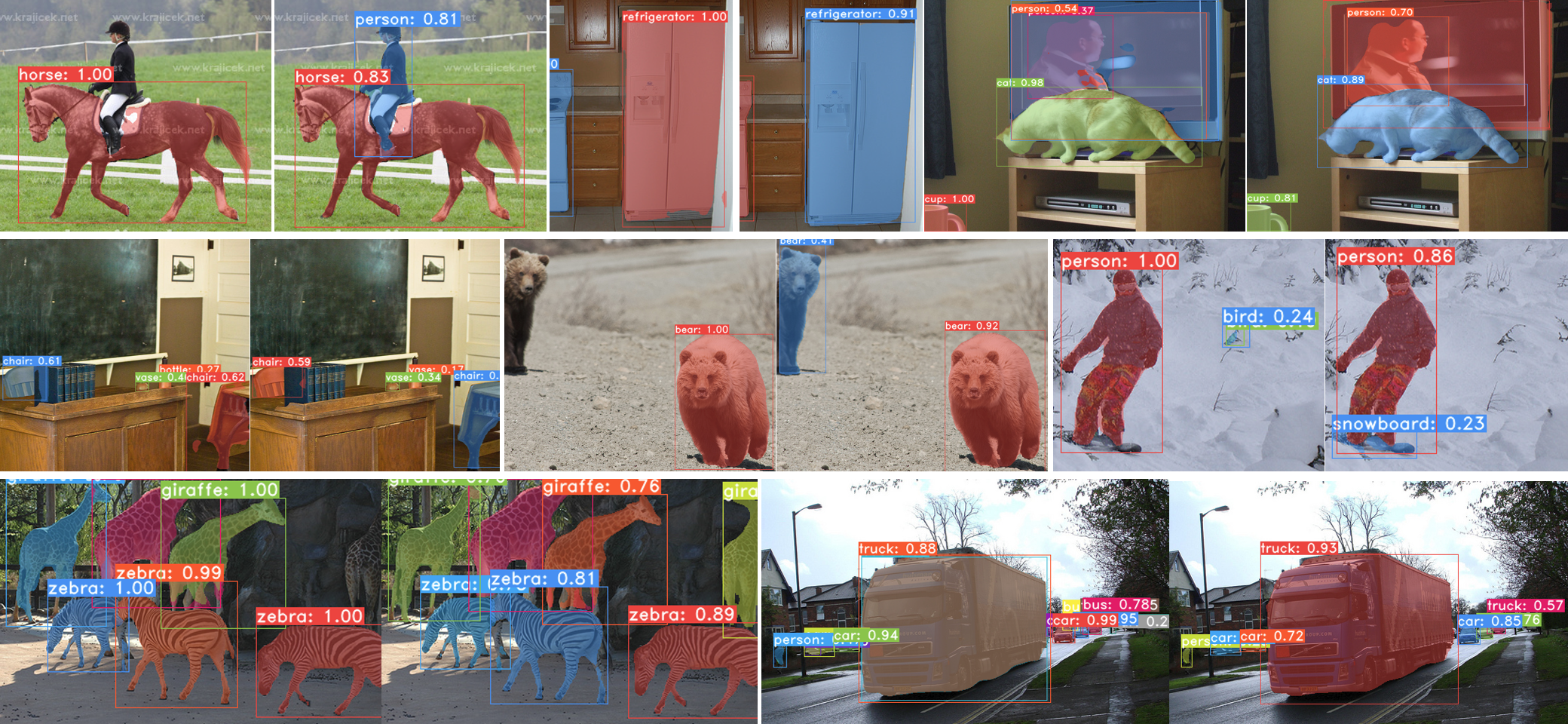}\\
\end{tabular}
\vspace{1mm}
    \caption[]{\textbf{Qualitative Comparisons between YOLACT \cite{bolya2019yolact} and \textit{B2Inst}-YOLACT.} 
    The image on the left is from the YOLACT while the image on the right is from ours.}
\label{fig:sup_yolact2ours}
\vspace{-5pt}
\end{figure*}

\begin{figure*}
\begin{tabular}{@{}c@{}}
\includegraphics[width=1\linewidth]{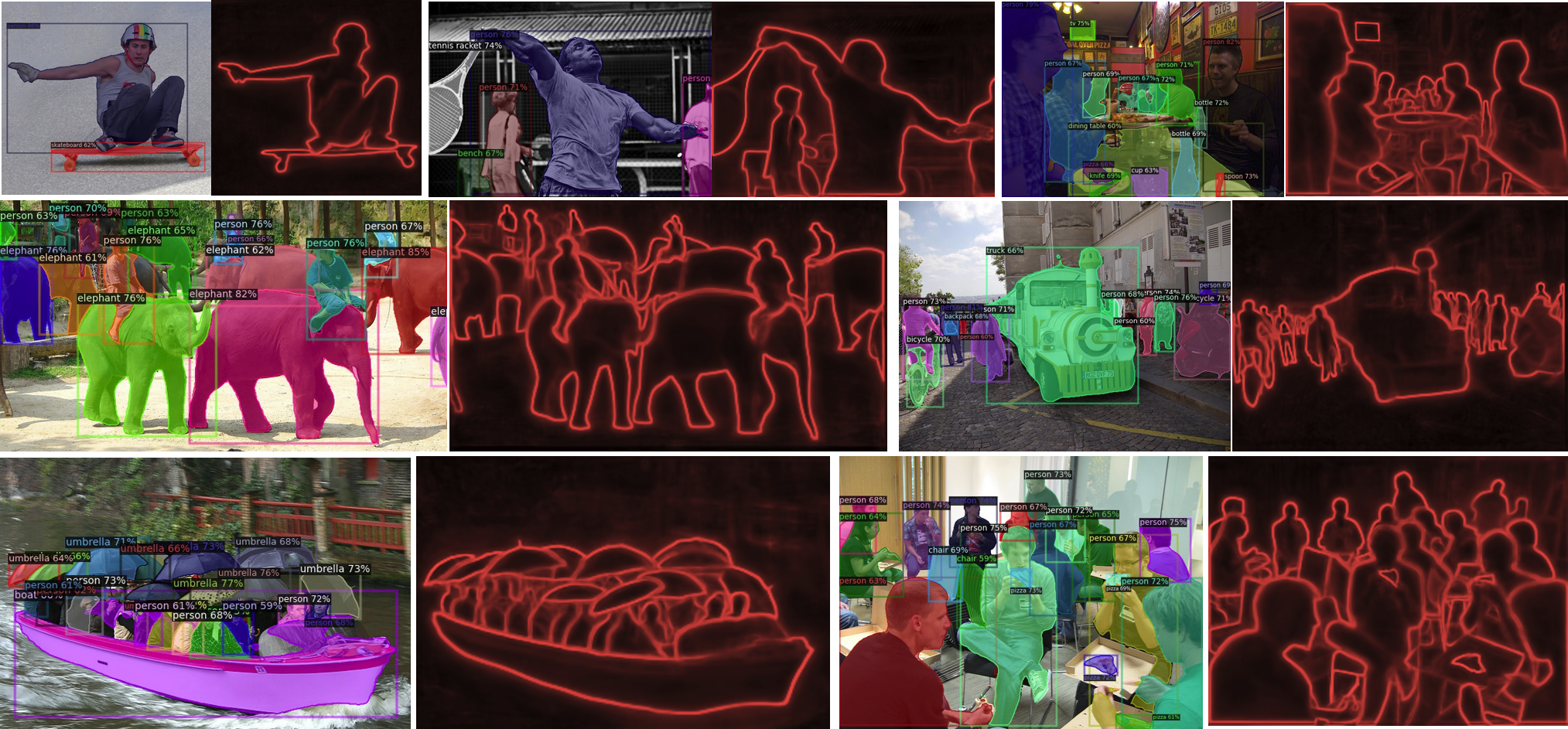}\\
\end{tabular}
\vspace{1mm}
    \caption[]{\textbf{The visualization of learned boundary basis.}
        }
\label{fig:sup_DIBwithHIB}
\vspace{-5pt}
\end{figure*}

\begin{figure*}
\begin{tabular}{@{}c@{}}
\includegraphics[width=1\linewidth]{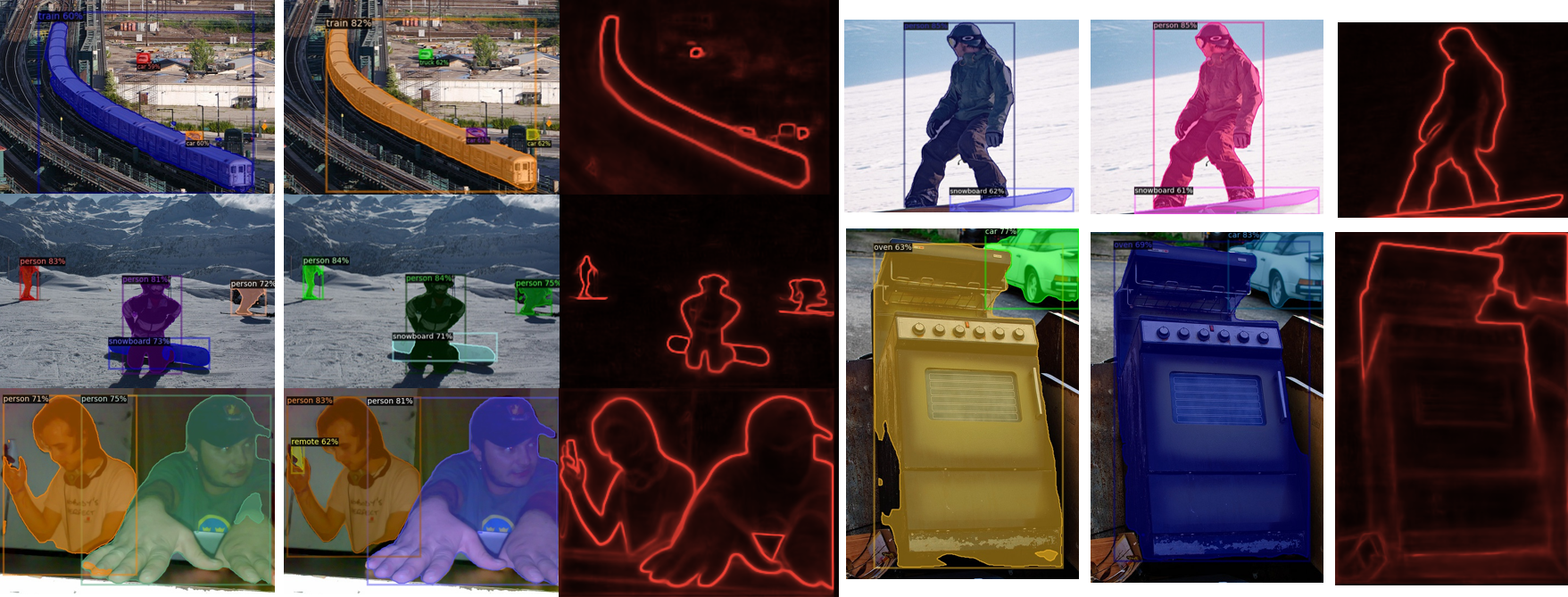}\\
\end{tabular}
\vspace{1mm}
    \caption[]{\textbf{Qualitative Comparisons on the Boundary Area.} (left) the results from BlendMask \cite{chen2020blendmask} (middle) the results from ours (right) the corresponding learned image boundary basis.}
\label{fig:sup_blend2ours2HIB}
\vspace{-5pt}
\end{figure*}

We provide additional qualitative results in \figref{fig:sup_blend2ours}, \figref{fig:sup_yolact2ours}, \figref{fig:sup_DIBwithHIB}, and \figref{fig:sup_blend2ours2HIB}.
\figref{fig:sup_blend2ours} shows the visual results of BlendMask~\cite{chen2020blendmask} and ours.
\figref{fig:sup_yolact2ours} shows the visual results of YOLACT~\cite{bolya2019yolact} and ours.
We can obviously observe that our method generates more accurate mask predictions.
Moreover, we provide learned boundary basis on \figref{fig:sup_DIBwithHIB}.
Finally, the superior boundary segmentation results over the baseline BlendMask are in \figref{fig:sup_blend2ours2HIB}.



\end{document}